\documentclass[letterpaper, 10 pt, conference]{ieeeconf}  

\IEEEoverridecommandlockouts                              

\usepackage{amsmath, amssymb}
\usepackage{graphicx}
\usepackage{url}
\usepackage{multirow}
\usepackage{xcolor}
\usepackage{hyperref}

\title{\LARGE \bf AIS-Based Vessel Trajectory Prediction Using Memory-Augmented Neural Networks}
\author{
  Wonmo Koo$^{1}$ \and Sanha Chang$^{1}$ \and Heeyoung Kim$^{1,\dagger}$%
  \thanks{$^{1}$Department of Industrial and Systems Engineering,
    Korea Advanced Institute of Science and Technology (KAIST),
    Daejeon, Republic of Korea.}%
  \thanks{$^\dagger$Corresponding author:
    \href{mailto:heeyoungkim@kaist.ac.kr}{heeyoungkim@kaist.ac.kr}}
}

\begin{document}

\maketitle
\thispagestyle{empty}
\pagestyle{empty}

\begin{abstract}
Accurate vessel trajectory prediction is essential for safe and efficient maritime operations, enabling collision avoidance and supporting route optimization. Although memory-augmented neural networks have recently shown strong performance in pedestrian and road-vehicle trajectory prediction by selectively retrieving relevant information from an external memory, their potential for vessel trajectory prediction remains underexplored. This paper presents an empirical investigation of memory-based trajectory prediction using Automatic Identification System (AIS) data. Experiments on data from the Gulf of Mexico and the New York Bight demonstrate consistent and substantial performance gains over a range of deep learning baselines that do not incorporate an external memory.
\end{abstract}

\section{INTRODUCTION}
Spatio-temporal data analysis has been widely studied across various domains due to its importance in understanding and predicting dynamic patterns over geographical areas \cite{chung2019crime, lee2020bayesian, koo2024deep}. Among its various applications, vessel trajectory prediction has gained increasing attention as a fundamental technology for maritime situational awareness, enabling collision avoidance and supporting route optimization \cite{kim2017early, zhang2022vessel}. As maritime traffic continues to grow in density and complexity, these capabilities are crucial for ensuring safe and efficient maritime operations, and have become even more important with the rapid development of autonomous navigation technologies \cite{xie2025ais}.

Recent studies predominantly rely on data from the Automatic Identification System (AIS) \cite{kim2017early, park2020maritime, zhang2022vessel, oh2023grid, xie2025ais, oh2025comparative}, which automatically identifies and tracks vessels through vessel-to-vessel and vessel-to-shore communication. Each AIS record typically includes vessel identifiers, such as Maritime Mobile Service Identity (MMSI) and call sign; dynamic navigational attributes, such as spatial position (latitude and longitude), speed over ground (SOG), and course over ground (COG); and static attributes, such as vessel type, length, and width. The mandatory installation of AIS for vessels above a certain gross tonnage under the International Maritime Organization (IMO) regulations has led to the availability of large-scale AIS data \cite{wolsing2022anomaly}. Consequently, a wide range of deep neural network-based approaches have been proposed for vessel trajectory prediction \cite{slaughter2025vessel, shin2024deep, xue2024g, jiang2024stmgf}.

When humans predict the future positions of moving agents (e.g., vessels, pedestrians, and road vehicles), they typically rely on reasoning informed by knowledge accumulated from past experiences and retrieved from memory. Inspired by this human behavior,
several approaches \cite{marchetti2020mantra, xu2022remember, marchetti2022explainable, yang2022continual, dong2023sparse} that leverage memory-augmented neural networks (MANNs) have been proposed, achieving remarkable performance across diverse trajectory prediction tasks. Such line of work commonly employs an external memory constructed from the training data in which each memory item stores a past-future encoding pair: the past encoding represents an observed trajectory segment, and the future encoding represents the associated future-side information. During prediction, they selectively retrieve relevant memory items by computing similarities between the current past encoding and stored past encodings. The retrieved future encodings are then decoded to provide future trajectory predictions.

Such memory-based approaches may be particularly appealing for vessel trajectory prediction because vessel motion patterns tend to be dominated by frequent cruising behaviors, whereas maneuvering patterns are more diverse but rare. By storing instance-level encodings, the external memory can explicitly retain individual trajectory instances, which may be beneficial for capturing both common cruising patterns and rare maneuvering behaviors \cite{santoro2016meta, choy2016looking,hyun2020memory,lee2023semi}.  However, existing memory-based approaches have been developed for pedestrians and road vehicles, and their potential for vessel trajectory prediction remains underexplored. 

In this paper, we provide empirical evidence that memory-based approaches can be highly effective for AIS-based vessel trajectory prediction. We adopt MANTRA \cite{marchetti2020mantra}, a seminal MANN for trajectory prediction. Unlike the original implementation, we incorporate SOG and COG values in addition to spatial positions as input features for memory writing and retrieval, given their importance in vessel trajectory prediction \cite{zhang2022vessel}. We use AIS data collected in the Gulf of Mexico and the New York Bight, and demonstrate the superior performance of the memory-based approach in terms of Average Displacement Error (ADE) and Final Displacement Error (FDE), compared to a variety of deep learning baselines that do not employ an external memory. The contributions of this work are summarized as follows:
\begin{itemize}
    \item To the best of our knowledge, we present the first empirical investigation of memory-based trajectory prediction in AIS-based vessel trajectory prediction. 
    \item Experiments on AIS data demonstrate that memory-based prediction can lead to substantial performance gains for vessel trajectory prediction.
    \item Across multiple prediction horizons, we reduced ADE and FDE by up to 46.4\% and 54.7\%, respectively, compared to the best-performing baseline on the Gulf of Mexico dataset (and by up to 33.3\% and 27.7\% on the New York Bight dataset).
\end{itemize}

\section{RELATED WORK}
\label{sec:related work}
\subsection{Memory-Based Trajectory Prediction}
Several memory-based approaches \cite{marchetti2020mantra, xu2022remember, marchetti2022explainable, yang2022continual, dong2023sparse} have been proposed for trajectory prediction. They typically leverage an external memory of past-future encoding pairs and a selective memory retrieval mechanism based on the similarity between the current past encoding and stored past encodings. However, they may differ in (i) what information is encoded in the future encodings and (ii) how similarity is computed for retrieval \cite{lee2013dependence,soh2018application}. Specifically, each future encoding may represent the full  trajectory \cite{marchetti2020mantra, yang2022continual}, a partially masked version of the trajectory \cite{dong2023sparse}, or only the trajectory endpoint \cite{xu2022remember}. Memory retrieval can be based on a fixed similarity measure (e.g., cosine similarity) \cite{marchetti2020mantra, yang2022continual, dong2023sparse} or a learnable  similarity function \cite{xu2022remember, marchetti2022explainable}. 
Despite the success of memory-based approaches in various trajectory prediction tasks, their effectiveness for vessel trajectory prediction remains underexplored.   

\subsection{Vessel Trajectory Prediction} 
In recent years, numerous deep neural network-based approaches have been proposed for vessel trajectory prediction \cite{zhang2022vessel, xie2025ais}. To capture temporal dependencies in vessel trajectories, several sequence modeling architectures have been widely adopted. Earlier approaches often rely on recurrent neural networks (RNNs) \cite{slaughter2025vessel, tang2022model} and temporal convolutional networks (TCNs) \cite{shin2024deep, lin2023ship}. More recently, Transformer-based models \cite{xue2024g, xiong2024informer} have gained popularity due to their ability to capture long-range dependencies through temporal self-attention mechanisms. In addition, spatio-temporal graph neural networks (ST-GNNs) \cite{jiang2024stmgf, liu2022stmgcn, wang2023novel} have been explored to explicitly model inter-vessel interactions. 

\section{METHODOLOGY}
We adopt MANTRA \cite{marchetti2020mantra}, a seminal memory-based trajectory prediction approach, with minor architectural modifications. In other words, the methodology described in this section closely follows the original MANTRA paper \cite{marchetti2020mantra}, except for the changes introduced to accommodate AIS data.

\label{sec:methodology}
\subsection{Problem Statement}
Let $t \in \{1, \dots, T_P + T_F\}$ be the time index, where $1\!:\!T_P$ denotes the observed history, and $T_P\!+\!1\!:\!T_P\!+\!T_F$ denotes the prediction horizon. For trajectory segment $i$, we denote the spatial position at time $t$ by 
\begin{equation*}
    y_t^i = (\phi_t^{i}, \lambda_t^{i}) \in \mathbb{R}^{2}, 
\end{equation*}
where $\phi_t^{i}$ and $\lambda_t^{i}$ are latitude and longitude (in degrees), respectively.

Given the past trajectory sequence $\textbf{y}_P^{i}$ and the associated covariates $\textbf{x}_P^{i}$,  our goal is to find a model $f$ that predicts the future trajectory sequence $\textbf{y}_F^{i}$. This can be formulated as follows: 
\begin{equation*}
    \hat{\textbf{y}}_F^{i} = f([\textbf{y}_{P}^{i}, \textbf{x}_P^{i}]),
\end{equation*}
where $\hat{\textbf{y}}_F^{i}\!=\!\{\hat{y}_{T_{P} + 1}^{i}, \dots, \hat{y}_{T_{P} + T_{F}}^{i}\}$ is the predicted trajectory for  $\textbf{y}_F^{i} \!=\! \{y_{T_{P} + 1}^{i}, \dots, y_{T_{P} + T_{F}}^{i}\}$, 
$\textbf{y}_P^{i} \!=\! \{y_{1}^{i}, \dots, y_{T_P}^{i}\}$, 
 $\textbf{x}_P^{i} \!=\! \{x_1^{i}, \dots, x_{T_P}^{i}\}$, and $[\cdot]$ denotes the concatenation operation. 

As SOG and COG are known to be important features for vessel trajectory prediction \cite{zhang2022vessel}, we incorporate them as the input covariates and define $x_t^i\in\textbf{x}_P^i$ as follows:
 \begin{equation*}
  x_{t}^{i} = (v_{t}^{i}, \cos(\psi_{t}^{i}), \sin(\psi_{t}^{i}), m_{t}^{i})\in\mathbb{R}^4,
\end{equation*}
where $v_t^{i}$ and $\psi_t^{i}$ are SOG (in knots) and COG (in radians) at $t$, respectively. Because AIS records may contain missing entries due to various reasons (e.g., limited coverage, communication outages, and sensor malfunctions), we apply imputation as a data preprocessing step to ensure that there are no missing values in $y_t^i$ and $x_t^i$ by independently interpolating each channel along the time axis.  Additionally, we incorporate a binary indicator $m_t^i$, where the value is 1 if $y_t^i$ is observed (and 0 otherwise), to inform the prediction model whether the position information at time $t$ was originally unobserved.

\subsection{Memory Construction and Model Training}
Using the training data consisting of $N$ trajectory segments, we construct an external memory through the following steps. 
First, we design an autoencoder consisting of two separate encoders, $g$ (past encoder) and $h$ (future encoder), and one decoder $d$. 
For each trajectory segment $i$, the past encoder takes $[\mathbf{y}_P^i, \mathbf{x}_P^i]$ as input, while the future encoder takes $[\mathbf{y}_F^i, \mathbf{x}_F^i]$ as input. 
The resulting past encoding $\pi^i$ and future encoding $\delta^i$ are then concatenated and fed into the decoder to reconstruct $\mathbf{y}_F^i$. Specifically, 
\begin{gather*}
    \pi^i = g([\textbf{y}_{P}^{i}, \textbf{x}_P^{i}]) \in \mathbb{R}^{d_{e}}, \quad \delta^i = h([\textbf{y}_{F}^{i}, \textbf{x}_F^{i}]) \in \mathbb{R}^{d_{e}},\\
    \tilde{\textbf{y}}_F^i = d([\pi^i, \delta^i]),
\end{gather*}
where $\tilde{\textbf{y}}_F^i\!=\!\{\tilde{y}_{T_P+1}^i, \dots, \tilde{y}_{T_P + T_F}^i\}$ denotes the reconstructed trajectory for $\textbf{y}_F^i$ and $\textbf{x}_F^i \!=\! \{x_{T_P+1}^{i}, \dots, x_{T_P+T_F}^i\}$. 

Second, the encoders and decoder are trained jointly by minimizing the reconstruction loss: 
\begin{equation*}
\label{eq:loss autoencoder}
    \mathcal{L}_{\text{ae}}
    = \frac{1}{NT_F}\sum_{i=1}^{N}\sum_{t=1}^{T_F}\|y_{T_P + t}^{i} - \tilde{y}_{T_P + t}^{i}\|_2^2,
\end{equation*}  
where $\|\cdot\|_2$ denotes the $\ell_2$-norm.
 
\begin{figure*}[t]
    \centering
    \includegraphics[width=0.75\textwidth]{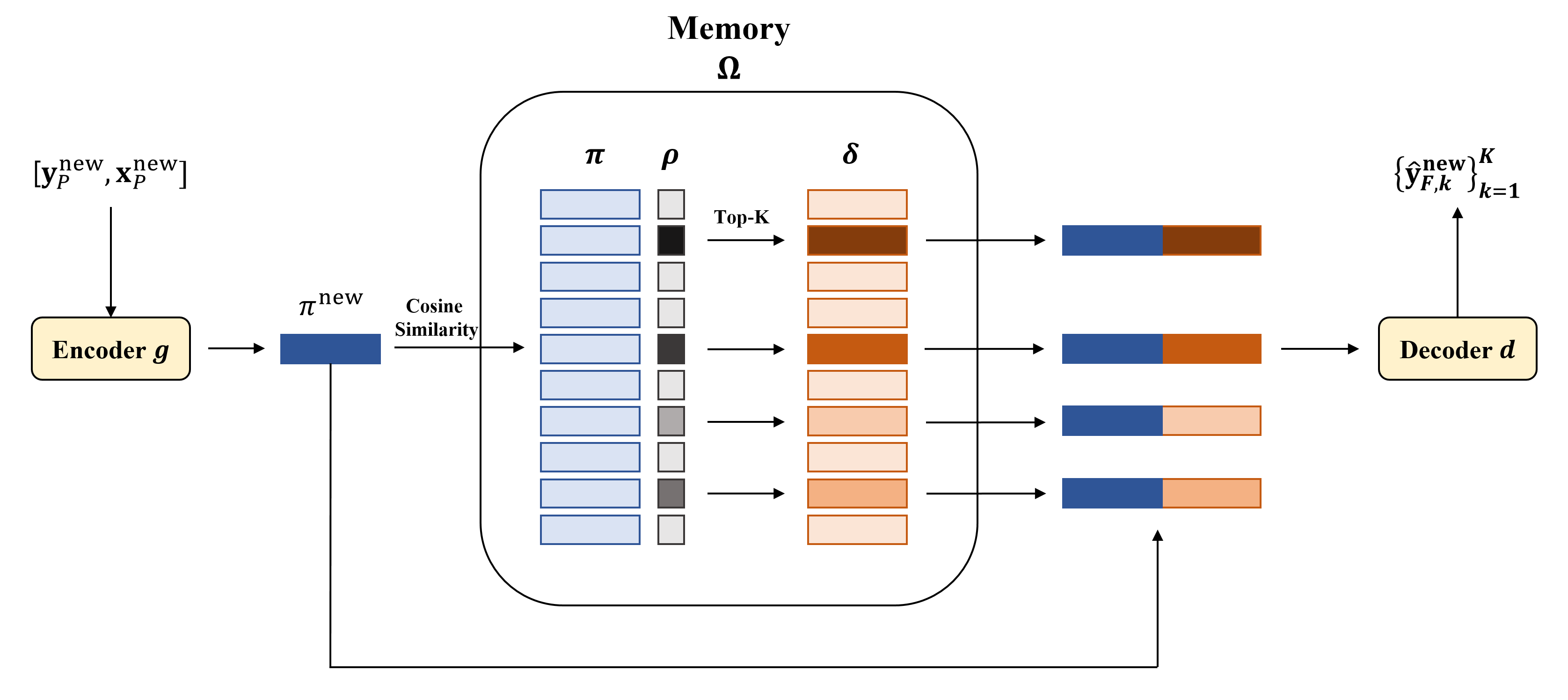}
    \caption{An overview of the memory-based trajectory prediction process.}
    \label{fig: prediction process}
\end{figure*}

Finally, we introduce a memory controller $c$ to construct a compact yet expressive external memory $\Omega$ from the set of $N$ past-future encoding pairs $\{(\pi^i, \delta^i)\}_{i=1}^{N}$ produced by the encoders $g$ and $h$ from $N$ training trajectory segments. The controller initializes the memory by randomly sampling a few encoding pairs and sequentially processes the training trajectory segments in a randomly shuffled order to decide whether each segment (its past-future encoding pair) should be written to the memory. To this end, the controller calculates a memory-write score $p^i \in (0,1)$, depending on the novelty of the target segment with respect to the current memory. The novelty of segment $i$ is quantified based on its future trajectory prediction error $e^i$, assuming the future part of segment $i$ is yet to be observed (the future prediction process will be described in Section \ref{subsec:memory based trajectory prediction}). The memory-write score $p^i$ is computed as follows:
\begin{gather*}
p^{i} = c(e^{i}) = \text{Sigmoid}(w_c \cdot e^{i}+b_c),\\
e^{i} = 1-\frac{1}{T_F}\sum_{t=1}^{T_F}\mathbf{1}\big(\|y_{T_P + t}^i - \hat{y}^{i}_{T_P+t}\|_2 < \epsilon\big),
\end{gather*}
where $w_c \in \mathbb{R}$ and $b_c \in \mathbb{R}$ are learnable weight and bias, respectively; $\mathbf{1}(\cdot)$ is an indicator function; $\epsilon$ is a hyperparameter. The controller writes $(\pi^i, \delta^i)$ into the memory if $p^{i} > 0.5$.

During controller training, only the controller parameters ($w_c$ and $b_c$) are updated (i.e., the encoders ($g$ and $h$) and the decoder $d$ remain fixed). At the start of each training epoch, the external memory is reinitialized, after which the memory is written sequentially while optimizing the controller with the following loss:
\begin{equation*}
    \mathcal{L}_c = e^{i}\cdot(1-p^i) + (1-e^i) \cdot p^i,
\end{equation*}
which encourages $p^i$ to increase with $e^i$. As a result, the controller is trained to interpret a high prediction error as evidence that the current memory lacks relevant memory items, and to write the corresponding trajectory segment into the memory.

\subsection{Memory-Based Trajectory Prediction}
\label{subsec:memory based trajectory prediction}
Given an external memory $\Omega = \{(\pi^{n_j}, \delta^{n_j})\}_{j=1}^{M}$, where $n_j\!\in\!\{1,\dots,N\}$ and $M \!<\! N$, we aim to predict the future trajectory $\textbf{y}_F^{\text{new}}$ when only $[\mathbf{y}_{P}^{\text{new}}, \mathbf{x}_{P}^{\text{new}}]$ is observed.
We compute the current past encoding $\pi^{\text{new}} = g([\mathbf{y}_{P}^{\text{new}}, \mathbf{x}_{P}^{\text{new}}])$ and retrieve relevant memory items from $\Omega$ based on cosine similarity between $\pi^{\text{new}}$ and $\pi^{n_j}$:
\begin{equation*}
    \rho^j = \frac{\pi^{\text{new}}\pi^{n_{j}^{'}}}{\|\pi^{\text{new}}\|_2 \cdot \|\pi^{n_j}\|_2}, \quad j=1,\dots M.
\end{equation*}

In this work, considering the multi-modal nature of vessel motions, we select the future encodings corresponding to the top-$K$ indices of $\rho^j$. Then, the selected future encodings are separately concatenated with $\pi^{\text{new}}$ and decoded into $K$ future trajectory predictions $\{\hat{\textbf{y}}_{F, k}^{\text{new}}\}_{k=1}^{K}$. We provide an illustration of the prediction process in Fig.~\ref{fig: prediction process}.

\section{EXPERIMENTS}
\label{sec:experiments}
\subsection{Dataset Description}
\begin{figure}[t]
    \centering
    \includegraphics[width=0.7\linewidth]{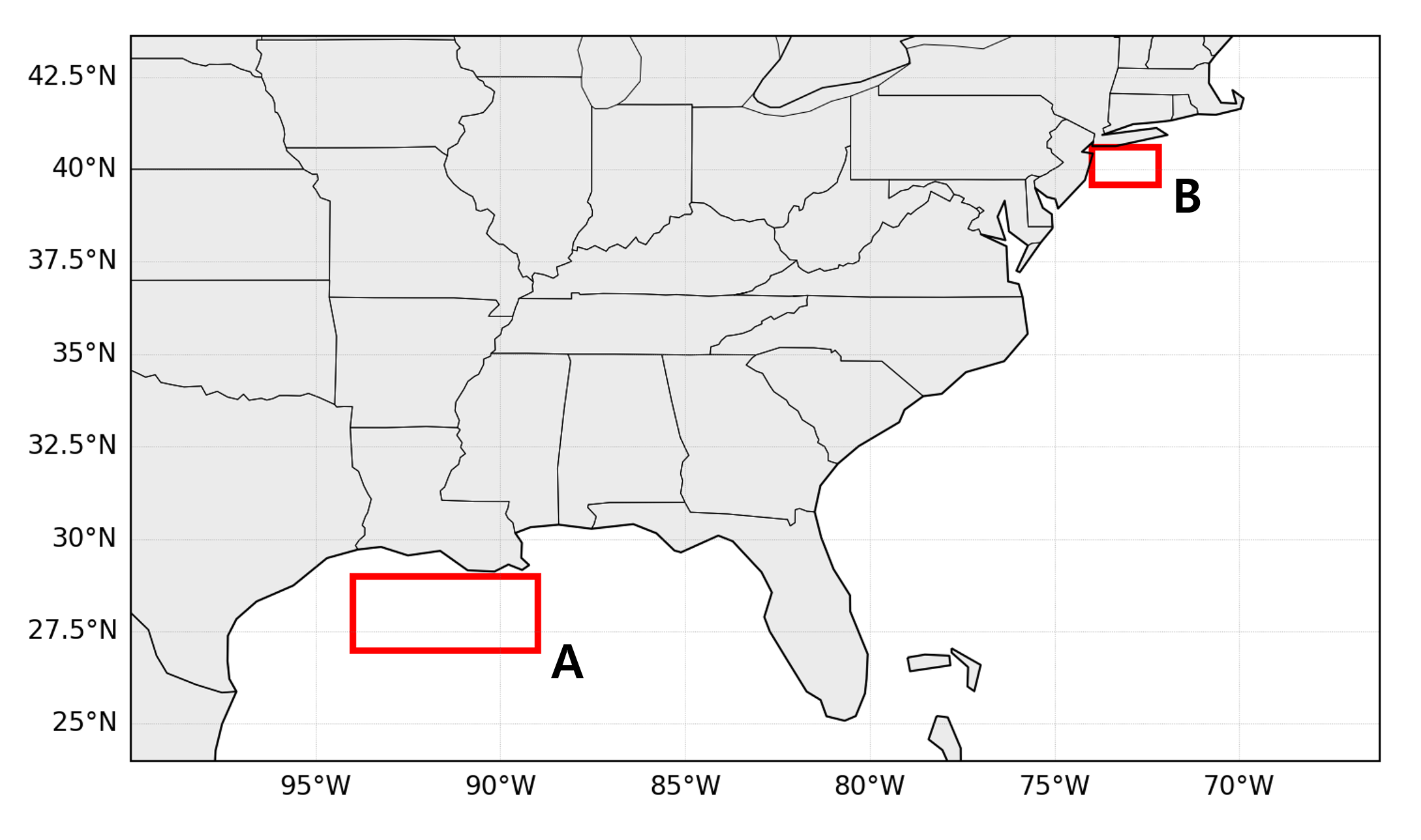}
    \caption{The red boxes indicate the study regions: (A) the Gulf of Mexico and (B) the New York Bight.}
    \label{fig:study_region}
\end{figure}

We used AIS data from the Gulf of Mexico and New York Bight collected by the National Oceanic and Atmospheric Administration (NOAA)\footnote{\url{https://hub.marinecadastre.gov}}
 from March 6 to March 8, 2023. The study regions were defined by latitude-longitude bounds of $[27.0^{\circ}, 29.0^{\circ}]$ and $[-94.0^{\circ}, -89.0^{\circ}]$ for the Gulf of Mexico, and $[39.6^{\circ}, 40.6^{\circ}]$ and $[-74.0^{\circ}, -72.2^{\circ}]$ for the New York Bight. Fig. \ref{fig:study_region} shows the selected regions, highlighted by red boxes. Only vessels with a length of at least 70 meters were considered. The resulting datasets contain 130,126 records from 369 vessels in the Gulf of Mexico and 22,741 records from 55 vessels in the New York Bight.

We removed records with SOG exceeding 30 knots, which were considered anomalous. The remaining data were resampled to a fixed 1-minute interval using mean aggregation. For each vessel, a continuous gap longer than 1 hour with no received records was regarded as the boundary between two voyages, while the remaining gaps ($\leq$ 1 hour) were treated as intra-voyage missing observations and imputed by applying cubic spline interpolation to spatial positions and linear interpolation to covariates. The voyages shorter than 1 hour were discarded. We performed a chronological split along the time axis, allocating the first 60\% of the time steps for training, the next 20\% for validation, and the remaining 20\% for testing. We applied $z$-score normalization to latitude, longitude, and SOG. 

\subsection{Evaluation Metrics}
To evaluate the prediction performance, we adopted two widely used metrics for trajectory prediction: ADE and FDE. ADE measures the average positional error over the prediction horizon, whereas FDE quantifies the error at the final destination.  

Following the evaluation protocol of prior work \cite{giuliari2021transformer, shi2023trajectory}, we calculated the minimum error over $K$ predicted trajectories to account for the intrinsic multi-modality of future vessel motions. As AIS data typically contain missing values, ADE was computed only over the observed time steps for which ground-truth positions are available, while FDE was evaluated at the last observed position. These metrics were computed as follows: 
\begin{align*}
    \text{ADE} &= \frac{1}{N}\sum_{i=1}^{N}\mathop{\min}\limits_{k \in \{1,\ldots,K\}}\bigg(\frac{1}{|S_F^i|}\sum_{t\in S_F^i}\|y^i_{T_P + t} - \hat{y}_{T_P+t, k}^i\|_{2}\bigg),\\
    \text{FDE} &=\frac{1}{N}\sum_{i=1}^{N}\mathop{\min}\limits_{k \in \{1,\ldots,K\}}||y^i_{T_P + t_{i}^*} - \hat{y}^i_{T_P + t_{i}^{*}, k}\|_{2},
\end{align*}
where $S_F^i = \{t\in\{1, \dots, T_F\}|m_{T_P + t}^i = 1\}$,   
$t_i^{*}$ denotes the last time step of $S_F^i$, and $\hat{y}_{T_P + t, k}^i$ is the $k$-th prediction result for $y_{T_P + t}^i$ among $K$ multi-modal predictions. 

\subsection{Baselines}
We compared MANTRA with six baselines that reflect recent research trends in vessel trajectory prediction, particularly the growing interest in multi-agent (vessel) interaction-aware models and Transformer-based architectures. Unlike most baselines---except TransformerTF---MANTRA does not explicitly model inter-agent interactions. Thus, improved performance would indicate that a memory-based approach can achieve competitive predictive capability even without explicit inter-agent interaction modeling.
\begin{itemize}
    \item STGAT \cite{huang2019stgat}: Spatio-Temporal Graph Attention Network (STGAT) is a sequence-to-sequence trajectory prediction model that employs a long short-term memory (LSTM) \cite{hochreiter1997long}-based encoder-decoder architecture. It models inter-agent interactions via a graph attention network \cite{velivckovic2017graph} and introduces random noise into the decoder input to enable multi-modal prediction. 
    \item Social-STGCNN \cite{mohamed2020social}: Social Spatio-Temporal Graph Convolutional Neural Network (Social-STGCNN) models trajectories of multiple agents as a dynamic spatio-temporal graph, where nodes represent agents and edges are weighted according to pairwise distances. It integrates spatial graph convolutions with temporal convolutions to capture inter-agent interactions and temporal motion dynamics, and provides bivariate Gaussian predictive distributions. 
    \item TransformerTF  \cite{giuliari2021transformer}: Transformer for Trajectory Forecasting (TransformerTF) employs a vanilla encoder-decoder Transformer \cite{vaswani2017attention} to capture long-range temporal dependencies in observed trajectories via temporal self-attention and autoregressively predicts future positions. It does not account for inter-agent interactions.
    \item AgentFormer \cite{yuan2021agentformer} : Agent-Aware Transformer (AgentFormer) is a Transformer-based trajectory prediction method that jointly models temporal dependencies and inter-agent interactions through a unified self-attention mechanism. For multi-modal prediction, it adopts a conditional variational autoencoder (CVAE)  framework \cite{kingma2013auto}, drawing multiple samples from the latent distribution to generate multiple plausible future trajectories.
    \item Social-Implicit \cite{mohamed2022social}: Social-Implicit is a multi-agent trajectory prediction model. It clusters agents based on the maximum change in their speed. For each cluster, spatial and temporal CNNs are employed to model inter-agent interactions and temporal motion dynamics, respectively. For multi-modal prediction, it injects random noise into the model input to generate multiple plausible future trajectories.
    \item TUTR \cite{shi2023trajectory}: Trajectory Unified Transformer (TUTR)  is an encoder-decoder Transformer  that models inter-agent interactions in the decoder and employs a dual-head structure to generate multi-modal trajectory predictions with associated mode probabilities.
\end{itemize}

\begin{table*}[tbp]
\caption{Performance comparison across different prediction horizons. The values in bold indicate the best scores, while underlined values represent the second-best scores. ADE and FDE are reported in units of $0.1^{\circ}$.}
\label{table:performance comparison}
\begin{center}
\resizebox{0.85\textwidth}{!}{%
\begin{tabular}{c|ccc|ccc}
\hline\hline
\multicolumn{7}{c}{\textbf{Gulf of Mexico}}\\
\hline
\multirow{2}{*}{\textbf{Method}} & \multicolumn{3}{c|}{\textbf{ADE}} & \multicolumn{3}{c}{\textbf{FDE}}\\
\cline{2-7}
&\textbf{10 minutes}&\textbf{20 minutes}&\textbf{30 minutes}&\textbf{10 minutes}&\textbf{20 minutes}&\textbf{30 minutes}\\\hline
STGAT &0.0644 (0.0089)&0.2172 (0.0510)&0.4276 (0.0229)&0.1320 (0.0232)&0.4495 (0.1194)&0.8787 (0.0264)\\
Social-STGCNN &1.1488 (0.0445)&1.5184 (0.1249) &1.8010 (0.0948)&1.2408 (0.0947)&1.6822 (0.1452)&2.0271 (0.1104)\\
TransformerTF &0.1300 (0.0129)&0.1480 (0.0165)&0.2628 (0.0810)&0.0882 (0.0081)&0.1196 (0.0209)&0.2398 (0.0629)\\
AgentFormer &0.4576 (0.0401)&0.7173 (0.1377)&0.9912 (0.3994)&0.4556 (0.0455)&0.7118 (0.1357)& 0.9856 (0.4290)\\
Social-Implicit &0.0292 (0.0061)&0.0532 (0.0068)&0.0847 (0.0128)&0.0468 (0.0116)&0.0965 (0.0131)&0.1616 (0.0243)\\
TUTR &\underline{0.0251} (0.0002)&\underline{0.0470} (0.0119)& \underline{0.0758} (0.0040)&\underline{0.0413} (0.0009)& \underline{0.0875} (0.0226)&\underline{0.1457} (0.0054)\\
\hline
MANTRA   &\textbf{0.0141} (0.0001)&\textbf{0.0252} (0.0004)& \textbf{0.0409} (0.0005)&\textbf{0.0187} (0.0004)&\textbf{0.0431} (0.0009)& \textbf{0.0771} (0.0008)\\
\hline\hline
\multicolumn{7}{c}{\textbf{New York Bight}}\\
\hline
STGAT &0.0644 (0.0089)&0.2172 (0.0510)&0.4276 (0.0229)&0.1320 (0.0232)&0.4495 (0.1194)&0.8787 (0.0264)\\
Social-STGCNN &1.0251 (0.0689)&1.5642 (0.0632)&1.9880 (0.1059)&1.0935 (0.0447)&1.7199 (0.0436)&2.2190 (0.1058)\\
TransformerTF &0.1086 (0.0222)&0.1278 (0.0292)&0.1970 (0.0381)&0.0884 (0.0210)&0.1245 (0.0326)&0.2313 (0.0658)\\
AgentFormer &0.3375 (0.0347)&0.4260 (0.0329)&0.4438 (0.0288)&0.3280 (0.0346)&0.4044 (0.0330)&0.4138 (0.0114)\\
Social-Implicit &\underline{0.0282} (0.0018)&\underline{0.0544} (0.0050)&\underline{0.0878} (0.0072)&\underline{0.0451} (0.0036)&\underline{0.0993} (0.0097)&\underline{0.1670} (0.0138)\\
TUTR &0.0724 (0.0019)&0.0784 (0.0059)&0.1252 (0.0062)&0.1226 (0.0096)&0.1469 (0.0075)&0.2453 (0.0122)\\
\hline
MANTRA   &\textbf{0.0188} (0.0011)&\textbf{0.0396} (0.0011)&\textbf{0.0638} (0.0006)&\textbf{0.0326} (0.0019)&\textbf{0.0735} (0.0018)&\textbf{0.1241} (0.0023)\\
\hline\hline
\end{tabular}}
\end{center}
\end{table*}

\subsection{Implementation Details}
We used the original implementation of MANTRA\footnote{\url{https://github.com/Marchetz/MANTRA-CVPR20}}. All components were trained using the Adam optimizer with a learning rate of $10^{-3}$ for up to 500 epochs. Early stopping with a patience of 20 was employed, and the batch size was set to 32 to approximate $\mathcal{L}_{\text{ae}}$. The hyperparameter for the memory controller $\epsilon$ was set to $3e^{-3}$. We used single-layer Gated Recurrent Unit (GRU) networks \cite{cho2014learning} with a hidden size of 96 for the two encoders and 192 for the decoder (i.e., $d_e = 96$).

We set $T_P \!=\! 30$ (corresponding to 30 minutes) and evaluated ADE and FDE over three prediction horizons, $T_F \!\in\! \{10, 20, 30\}$, with $K\!=\!5$.
We performed sliding-window segmentation with a window length of $T_P + T_F$ time steps and a stride of 5.
When $T_F = 30$, this produced 18,259/6,398/5,529 training/validation/test segments for the Gulf of Mexico and 2,799/715/952 segments for the New York Bight (shorter $T_F$ yielded slightly larger sample sizes).

\subsection{Quantitative Analysis}
Table \ref{table:performance comparison} shows the results over five independent runs, with the average values and standard errors in parentheses. At all prediction horizons, MANTRA outperformed all baselines by a large margin in both ADE and FDE, despite the absence of explicit inter-vessel interaction modeling. Specifically, in the Gulf of Mexico, compared to the best-performing baseline, MANTRA improved ADE by $\mathbf{43.8\%/46.4\%/46.0\%}$ and FDE by $\mathbf{54.7\%/50.7\%/47.1\%}$ for prediction horizons of 10/20/30, respectively. In the New York Bight, the corresponding improvements were $\mathbf{33.3\%/28.5\%/27.3\%}$ in ADE and $\mathbf{27.7\%/26.0\%/25.7\%}$ in FDE. Moreover, MANTRA consistently exhibited the smallest standard errors across all settings, indicating more stable performance over random seeds compared to the baselines. These findings may highlight the potential of memory-based prediction as a powerful paradigm for vessel trajectory prediction.

\subsection{Qualitative Analysis}
\begin{figure}[h!]
\centering
\includegraphics[width=0.65\linewidth]{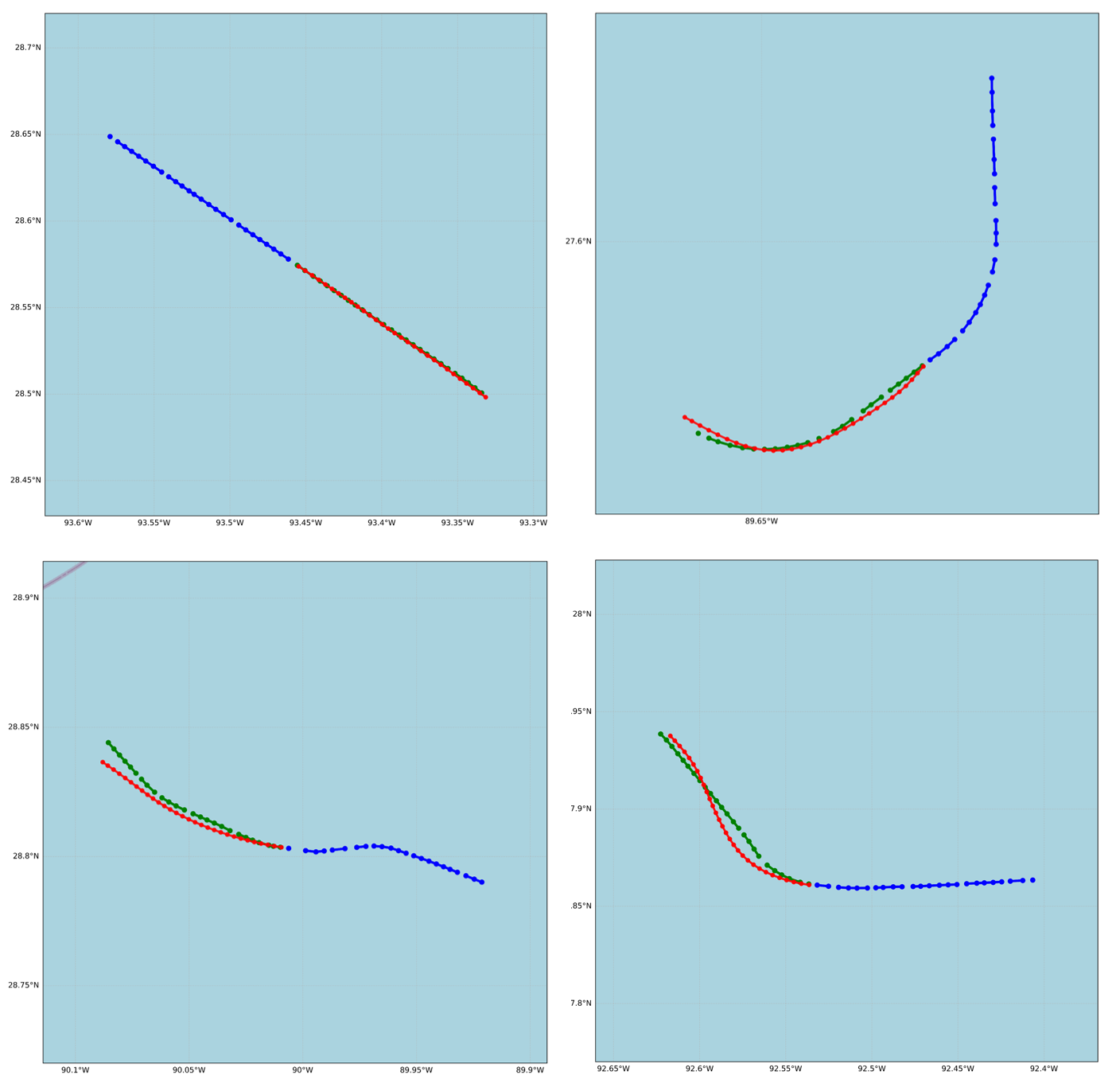}
\caption{Examples of trajectory prediction results on the Gulf of Mexico with a 30-minute observation window and a 30-minute prediction horizon. In each panel, blue markers indicate the past trajectory, green markers indicate the ground-truth future trajectory, and red markers indicate the predicted trajectories. Gaps between markers in the past and ground-truth future trajectories indicate intervals with missing observations.}
\label{fig:qualitative}
\vspace{-0.3 in}
\end{figure}

In Fig.~\ref{fig:qualitative}, we visualize the best-of-5 trajectory predictions of MANTRA (with the lowest ADE) on the Gulf of Mexico dataset for a 30-minute prediction horizon ($T_F\!=\!30$). In each panel, blue markers indicate the past trajectory, green markers denote the ground-truth future trajectory, and red markers represent the predicted trajectory. 
For the past and ground-truth future trajectories, gaps between markers indicate intervals with missing observations. 

As shown in the figure, MANTRA closely matches the ground-truth future trajectories for both cruising (upper-left panel) and maneuvering cases (remaining panels).

\section{CONCLUSION}
In this study, we empirically investigated the effectiveness of memory-based trajectory prediction for AIS-based vessel trajectory prediction. Experiments demonstrated substantial improvements over a range of deep learning baselines that do not incorporate an external memory. By explicitly storing instance-level trajectory representations, the memory-based approach (MANTRA) was able to capture both common cruising patterns and rare maneuvering behaviors of maritime vessels. These findings suggest that memory-based approaches may provide a promising direction for modeling complex and diverse vessel movement patterns and may hold significant potential for improving vessel trajectory prediction systems. 
Future work will incorporate high-dimensional navigational contextual variables, such as vessel type and size, weather conditions, and waterway geometry, that may significantly affect vessel operations, following the approach in \cite{kim2023contextual}. In addition, we will investigate methods for quantifying prediction uncertainty \cite{kim2021locally,yoon2024uncertainty}.
\color{black}




\section{ACKNOWLEDGMENT}
This work was supported by the National Research Foundation of Korea (NRF) grant funded by the Korea government (MSIT) (2023R1A2C2005453, RS-2023-00218913).

\bibliographystyle{IEEEtran}
\bibliography{references}

\end{document}